\documentclass{article}

\usepackage[final]{neurips_2020}

\usepackage[utf8]{inputenc} 
\usepackage[T1]{fontenc}    
\usepackage{url}            
\usepackage{booktabs}       
\usepackage{amsfonts}       
\usepackage{nicefrac}       
\usepackage{microtype}      
\usepackage{xcolor}         

\usepackage{enumitem}
\usepackage{graphicx}
\usepackage{multirow}
\usepackage{amsmath,amssymb} 
\usepackage[colorlinks,citecolor=blue,urlcolor=magenta,bookmarks=false,hypertexnames=true]{hyperref}

\title{DroneNet: Crowd Density Estimation using Self-ONNs for Drones}

\author{%
  Muhammad Asif Khan\thanks{Corresponding author} \\
  Qatar Mobility Innovations Center (QMIC)\\
  Qatar University\\
  Doha, Qatar \\
  \texttt{mkhan@qu.edu.qa} \\
  \And
  Hamid Menouar \\
  Qatar Mobility Innovations Center (QMIC) \\
  Qatar University\\
  Doha, Qatar \\
  \texttt{hamidm@qmic.com} \\
  \And
  Ridha Hamila \\
  Department of Electrical Engineering \\
  Qatar University\\
  Doha, Qatar \\
  \texttt{hamila@qu.edu.qa} \\
}

\begin{document}
\maketitle
\begin{abstract}
Video surveillance using drones is both convenient and efficient due to the ease of deployment and unobstructed movement of drones in many scenarios. An interesting application of drone-based video surveillance is to estimate crowd density (both pedestrians and vehicles) in public places. Deep learning using convolution neural networks (CNNs) is employed for automatic crowd counting and density estimation using images and videos. However, the performance and accuracy of such models typically depends upon the model architecture i.e., deeper CNN models improve accuracy at the cost of increased inference time. In this paper, we propose a novel crowd density estimation model for drones (DroneNet) using Self-organized Operational Neural Networks (Self-ONN). Self-ONN provides efficient learning capabilities with lower computational complexity as compared to CNN-based models. We tested our algorithm on two drone-view public datasets. Our evaluation shows that the proposed DroneNet shows superior performance on an equivalent CNN-based model.
\end{abstract}
\section{Introduction} \label{sec:intro}
Automatic aerial video surveillance using drones has many potential applications in future smart cities. Due to the ease of deployment, and flexibility to reach anywhere, drones can play a vital role in several scenarios such as crowd surveillance for smart policing in public places e.g., metro stations, stadiums, and political rallies. \cite{Simpson2021, khan_1}, situational awareness during disasters \cite{Sambolek2021}, traffic monitoring in smart transportation systems \cite{Sun2022}, and monitoring of forests and wild life \cite{Chandana_2022, Allauddin2019}.
Traditionally crowd counting using images employs handcrafted local features such as full body \cite{Topkaya_2014, Tuzel_2008}, body parts \cite{Li_2008, Felzenszwalb_2010}, shapes \cite{Lin_2010}, or global features such as texture \cite{Chen_2012}, edges \cite{Wu_2006}, foreground \cite{Davies_1995} and gradients \cite{Dalal_2005, Tian_2010} to detect and count people. These methods perform poorly on images of dense crowds with severe occlusions and other variations \cite{Khan2022RevisitingCC}. To overcome these challenges, CNN-based crowd counting has been introduced in \cite{CrowdCNN_CVPR2015, CrowdNet_CVPR2016}. Several state-of-the-art CNN models are developed over time mainly to improve the accuracy in more challenging scenes \cite{MCNN_CVPR2016, MSCNN_ICIP2017, CSRNet_CVPR2018, TEDnet_CVPR2019, SANet_ECCV2018, SASNet_AAAI2021}. However, while achieving higher accuracy, the complexity of the model is often ignored. Thus, several of these models use popular deep CNN networks as a frontend network to extract fine-grained features. For instance, VGG-16 \cite{VGG16_ICLR2015} is used as a front-end in \cite{CrowdNet_CVPR2016, CSRNet_CVPR2018, CANNet_CVPR2019}, ResNet \cite{ResNet_CVPR2016} is used in \cite{MMCNN_ACCV2020, MFCC_2022}, and Inception \cite{Inception_CVPR2015} in \cite{SGANet_IEEEITS2022}. Such deep CNN architectures are better suited for applications using powerful GPU servers, but when run on edge devices with low computation power, they incur higher inference delays. Thus, lightweight CNN models are desired for edge-based processing. However, the lower accuracy of lightweight models often precludes the adoption of such models in practical applications.
\par
Recently, authors in \cite{ONN_2020} introduces Operational Neural Networks (ONNs) architectures to achieve higher accuracy with small models consisting of fewer layers. ONNs replace the non-linear convolution operation in neurons with a set of non-linear operations to improve the learning process in more challenging tasks. The Self-organized ONNs (which is an improved version of ONNs) \cite{SelfONN_imgrestore_2021} with even compact architecture show superior performance over conventional CNNs of equivalent or larger sizes in several problems e.g., image restoration \cite{ONN_2020, SelfONN_imgrestore_2021}, video segmentation \cite{hamila_1}, medical imaging \cite{hamila_2}. Inspired by the improved results of SelfONNs in various tasks, we adopted the SelfONN paradigm in our proposed crowd density estimation model.
\par
The contribution of the paper is as follows: We propose a compact crowd counting and density estimation model using Self-ONN \cite{SelfONN_imgrestore_2021} architecture. To our knowledge, the use of SelfONN in crowd density estimation has not been proposed earlier. The proposed model is tested on benchmark crowd datasets containing images taken with drones. The performance is compared with an equivalent CNN model as well as other existing state-of-the-art crowd counting models over various metrics to show significant performance improvement.

\section{Related Work} \label{sec:rel_work}
CNN-based crowd counting and density estimation was first proposed in \cite{CrowdCNN_CVPR2015} using a single-column CNN network that consists of six layers. Following the success of CNNs, several other works proposed different CNN architectures in attempts to achieve better accuracy over benchmark datasets. The major architectural changes introduced over time include multi-column networks \cite{MCNN_CVPR2016, CrowdNet_CVPR2016, SCNN_CVPR2017, CMTL_AVSS2017}, modular networks \cite{MSCNN_ICIP2017, SANet_ECCV2018, SGANet_IEEEITS2022}, encoder-decoder models \cite{TEDnet_CVPR2019, MobileCount_PRCV2019}, and models using transfer learning \cite{CSRNet_CVPR2018, CANNet_CVPR2019, GSP_CVPR2019, TAFNET_2022}.
Small-sized models with single-column architectures (e.g. \cite{CrowdCNN_CVPR2015} typically suffers from low accuracy when the images have scale variations. Scale variations arise from the camera perspective distortions i.e., objects closer to the camera are bigger than those far from the camera. Thus, multi-column networks with filters of different receptive fields in each column are used to capture these scale variations. For instance, the multi-column CNN (MCNN) proposed in \cite{MCNN_CVPR2016} uses a three-column architecture with filters of variable sizes ($9\times9, 7\times7, 5\times5, \text{and } 3\times3$) in different columns. The switching-CNN \cite{SCNN_CVPR2017} uses three CNN networks (regressor networks) dynamically selected by another CNN network (classifier or switch). The input image is thus passed through only one column (regressor) based on the image crowd density determined by the classifier. A drawback of multi-column CNNs is that their capability to adapt to scale variations is limited by the number of columns. Thus, the model size will significantly increase when there are large-scale variations in the dataset. An alternative solution is to use modular networks which use single-column architecture with special scale-adaptive modules. These models are inspired by the Inception model \cite{Inception_CVPR2015}. Another category of crowd counting models is the encoder-decoder models \cite{TEDnet_CVPR2019, MobileCount_PRCV2019} inspired from the UNet architecture \cite{UNet_2015}, where an encoder network learns and extracts features from the network and a decoder network then uses these features to predict the density map. Encoder-decoder models are good when high-quality density maps are desired. Crowd counting can become more challenging when applied to very dense and congested scenes. Thus, a large number of research works propose the use of transfer learning i.e., a pretrained image classification model such as VGG \cite{VGG16_ICLR2015}, ResNet \cite{ResNet_CVPR2016} or Inception \cite{Inception_CVPR2015} as a front-end to extract features and then a small CNN network uses these features to estimate the crowd density. Transfer-learning based approaches are generally more accurate and faster to train. However, these models incur longer inference delays and require more memory to store and execute.

\section{Proposed Scheme} \label{sec:scheme}

In aerial crowd surveillance applications using drones, we aim at designing a lightweight architecture that can run faster on low-end processors and provide sufficient accuracy. We developed a lightweight model following a similar architecture of MCNN \cite{MCNN_CVPR2016} but replacing the convolution layers with the SelfONN layers. To keep the model size compact for the intended application, we do not use transfer learning.

\subsection{The DroneNet Architecture}
\begin{figure*}[ht]
    \centering
    \includegraphics[width=0.9\textwidth]{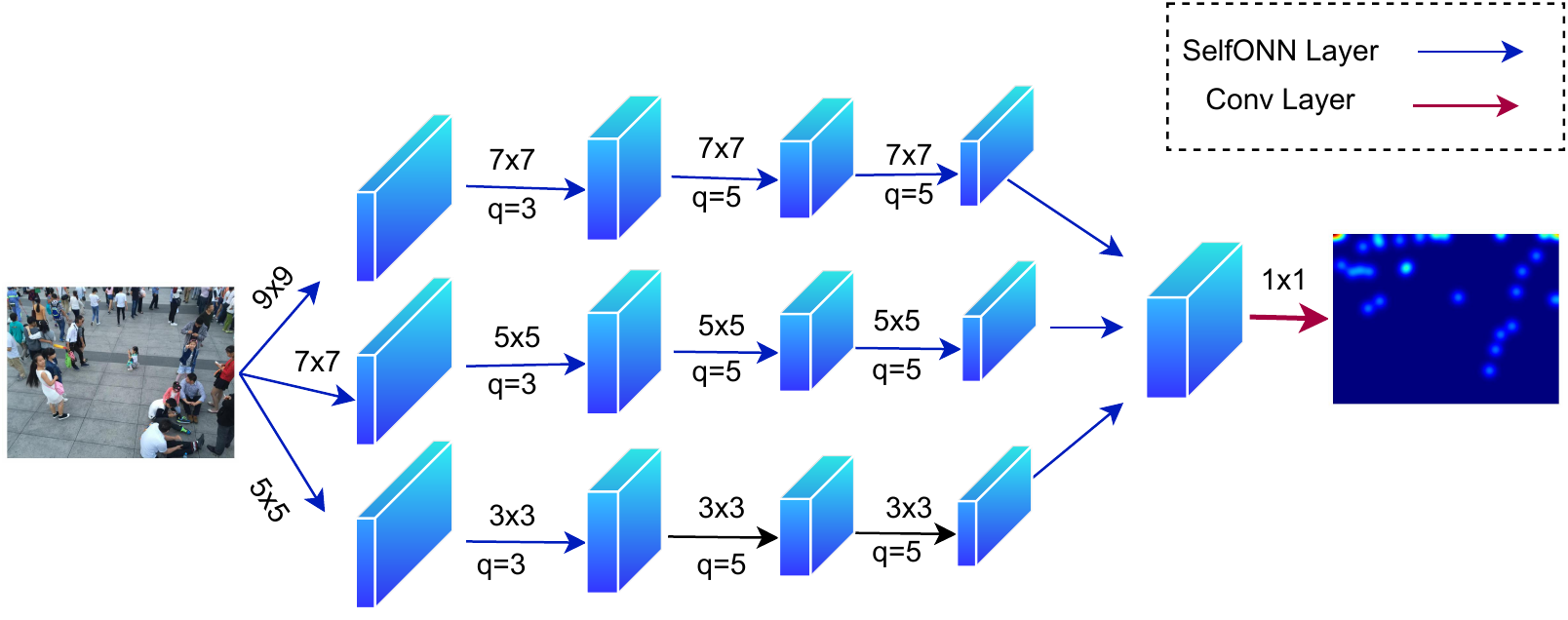}
    \caption{Architecture of DroneNet.}
    \label{fig:DroneNet}
\end{figure*}

The architecture of our proposed network (DroneNet) is shown in Fig. \ref{fig:DroneNet}. Like MCNN, it is a three-column CNN architecture with the same number of layers in each column. The difference is that all convolution layers in the three columns are replaced with SelfONN layers except the last ($1\times1$) convolution layer after columns concatenation. We used Tanh activation layers after each SelfONN layer except the last convolution layer which is proceeded by a Relu activation layer.

\subsection{Hyper-parameters Settings}
A SelfONN layer uses an additional hyper-parameter $q$ which introduces non-linearity in the neurons. We set the value of $q=3$ in the first layer of each column, whereas all the remaining SelfONN layers use $q=5$. The last standard convolution layer does not use $q$.

\subsection{Model Training}
The DroneNet model is trained over the DroneRGBT dataset \cite{DroneRGBT_dataset}. The DroneRGBT dataset has 1807 RGB and thermal image pairs in the train set. Each image has a spatial resolution of 512×640 pixels. We split the train set into a ratio of ($70\%:30\%$) for training and validation. The dataset covers several scenes (e.g., campus, streets, public parks, car parking, stadiums, and plazas) and contains diverse crowd densities, illumination, and scale variations. The dataset provides the head locations of people (called dot annotations). To train the model, the dot annotations are converted to density maps that serve as ground truth for the images. The density map is generated by convolving a delta function $\delta(x - x_i)$ with a Gaussian kernel $G_\sigma$, where $x_i$ is a pixel containing the head position.
\begin{equation}
    D = \sum_{i=1}^{N}{ \delta(x-x_i) * G_\sigma}
\end{equation}
where, $N$ is the total number of annotated points (i.e., the total count of heads) in the image. We empirically determined a fixed value of $\sigma=7$ that provides a good estimation of the head sizes. We employ full image-based training instead of patch-based training for simplicity and speed. To avoid model overfitting, data augmentation techniques including horizontal flipping, and random brightness and contrast are applied. We use Adam optimizer \cite{Adam_ICLR2015} with a base learning rate of $0.0001$. The loss function used is pixel-wise Euclidean distance between the target and predicted density maps which are defined in Eq. \ref{eq:mse}.

\begin{equation} \label{eq:mse}
    L(\Theta) = \frac{1}{N} \sum_{1}^{N}{ ||D(X_i;\Theta) - D_i^{gt}||_2^2}
\end{equation}

where $N$ is the number of samples in training data, $D(X_i;\Theta)$ is the predicted density map with parameters $\Theta$ for the input image $X_i$, and $D_i^{gt}$ is the ground truth density map. The model was trained on two GPUs (Nvidia RTX-8000) using PyTorch deep learning framework.

\section{Evaluation and Results} \label{sec:results}

\subsection{Evaluation Metrics} \label{subsec:metrics}
We evaluated the performance of DroneNet using eight (8) metrics including mean absolute error (MAE) and Grid Average Mean Error (GAME), Structural Similarity Index (SSIM), Peak Signal-to-Noise Ratio (PSNR), model size (MB), Giga Multiply-Accumulate operations (GMACs), inference time (mili-second), and throughput (frames per second).

MAE and GAME provide the accuracy of the model and are calculated as \ref{eq:mae} and \ref{eq:game}, respectively:

\begin{equation} \label{eq:mae}
    MAE = \frac{1}{N} \sum_{1}^{N}{(e_n - \hat{g_n})}
\end{equation}

where, $N$ is the total number of images in the dataset, $g_n$ is the ground truth (actual count) and $\hat{e_n}$ is the prediction (estimated count) in the $n^{th}$ image.

\begin{equation} \label{eq:game}
    GAME = \frac{1}{N} \sum_{n=1}^{N}{ ( \sum_{l=1}^{4^L}{|e_n^l - g_n^l|)}}
\end{equation}
We set the value of $L=4$, thus each density map is divided into a grid size of $4\times4$ creating $16$ patches.

The SSIM and PSNR metrics measure the quality of predicted density maps as compared to ground truth density maps and are calculated as follows \ref{eq:ssim} \ref{eq:psnr}:

\begin{equation} \label{eq:ssim}
    SSIM (x,y) = \frac{(2\mu_x \mu_y + C_1)  (2\sigma_x \sigma_y C_2)}  {(\mu_z^2 \mu_y^2 + C_1)  (\mu_z^2 \mu_y^2 + C_2)}
\end{equation}
where $\mu_x, \mu_y, \sigma_x, \sigma_y$ represent the means and standard deviations of the actual and predicted density maps, respectively.

\begin{equation} \label{eq:psnr}
    PSNR = 10 log_{10}\left( \frac{Max(I^2)}{MSE}  \right)
\end{equation}
where $Max(I^2)$ is the maximal in the image data. If it is an 8-bit unsigned integer data type, the $Max(I^2)=255$.

The other metrics such as the size of the model can be useful to know the required storage space (especially when a device does not have external memory and has limited on-chip memory), GMACs can be an indication of the model time complexity as well as can be used to estimate the model energy consumption. Similarly, the inference time and throughput are both related to the execution speed and can be used in application configuration (e.g., frame capturing rate of the camera, etc.).

\subsection{Results} \label{subsec:results}
\begin{figure*}[!h]
    \centering
    \includegraphics[width=0.99\textwidth]{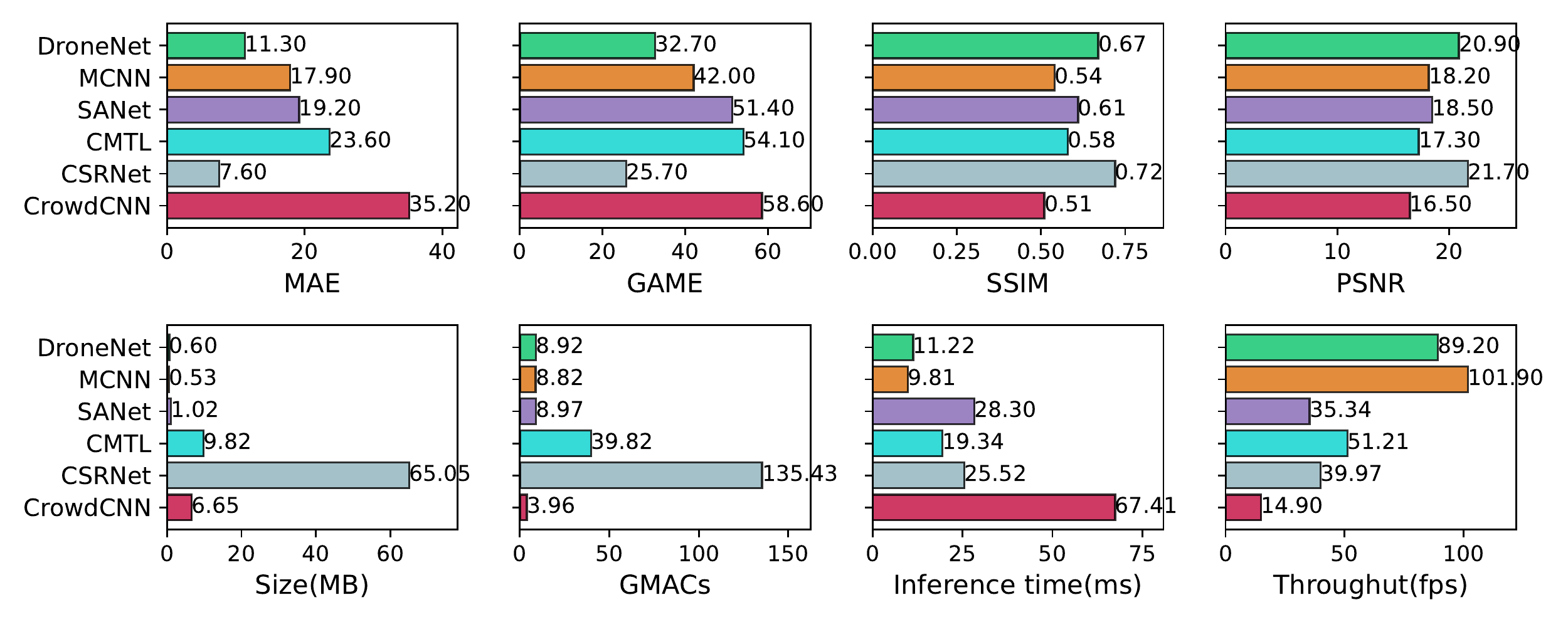}
    \caption{Performance evaluation of DroneNet over various metrics.}
    \label{fig:comparison}
\end{figure*}

The DroneNet is trained using the aforementioned method and hyper-parameter settings and evaluated over various metrics. The performance of the DroneNet is compared over eight (8) metrics against five (5) other crowd counting models namely CrowdCNN \cite{CrowdCNN_CVPR2015}, MCNN \cite{MCNN_CVPR2016}, CMTL \cite{CMTL_AVSS2017}, CSRNet \cite{CSRNet_CVPR2018}, and SANet \cite{SANet_ECCV2018}. The results are graphically illustrated in Fig. \ref{fig:comparison}.

The analysis shows that DroneNet achieves better accuracy (MAE and GAME) than CrowdCNN, MCNN, CMTL, and SANet and is very close to that CSRNet. Importantly, the DroneNet using Self-ONN achieves higher accuracy than the equivalent CNN-based model (MCNN). In addition, DroneNet also achieves equal or slightly better SSIM and PSNR values than MCNN.
In terms of model size, DroneNet has a higher size than MCNN. However, the size is still of the order that can be easily stored in on-chip memory as compared to the deep models such as CSRNet (which is about $25\times$ higher). The inference-time comparison also shows that DroneNet achieves a throughput closer to that of MCNN and much higher than all other models.

\begin{figure*}[!h]
    \centering
    \includegraphics[width=0.9\textwidth]{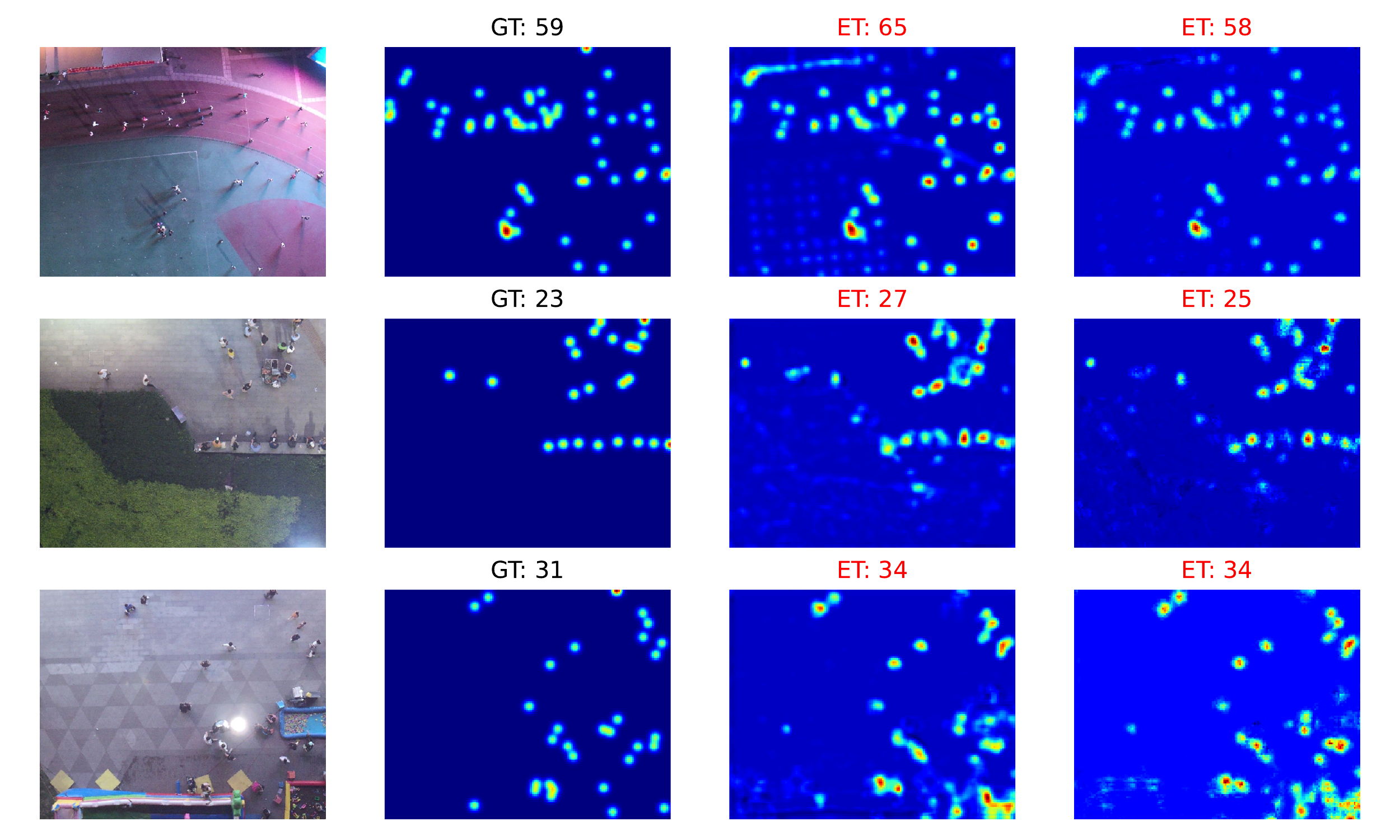}
    \caption{Sample Predictions: The first column shows images from the test set. The second column shows the ground truth for the given images. The third and fourth columns show predictions using MCNN \cite{MCNN_CVPR2016} and proposed DroneNet.}
    \label{fig:predictions}
\end{figure*}

Fig. \ref{fig:predictions} shows sample predictions using DroneNet and other CNN-based networks.

\subsection{Ablation Study} \label{subsec:ablation}
To validate the accuracy of DroneNet, we further investigated its performance over two other benchmark datasets used in crowd counting studies i.e., ShanghaiTech Part-B \cite{MCNN_CVPR2016} and CARPK \cite{CARPK_dataset}. The trained DroneNet is trained using the same hyper-parameter settings with different values of $\sigma$ for ground truth density-map generation. We use $\sigma=15$ for both datasets as in \cite{CSRNet_CVPR2018, BL_ICCV2019}. Our evaluation shows that DroneNet achieves better accuracy (low MAE and GAME values) than MCNN over both datasets \cite{MCNN_CVPR2016}.

\begin{figure}[!h]
    \centering
    \includegraphics[width=0.7\columnwidth]{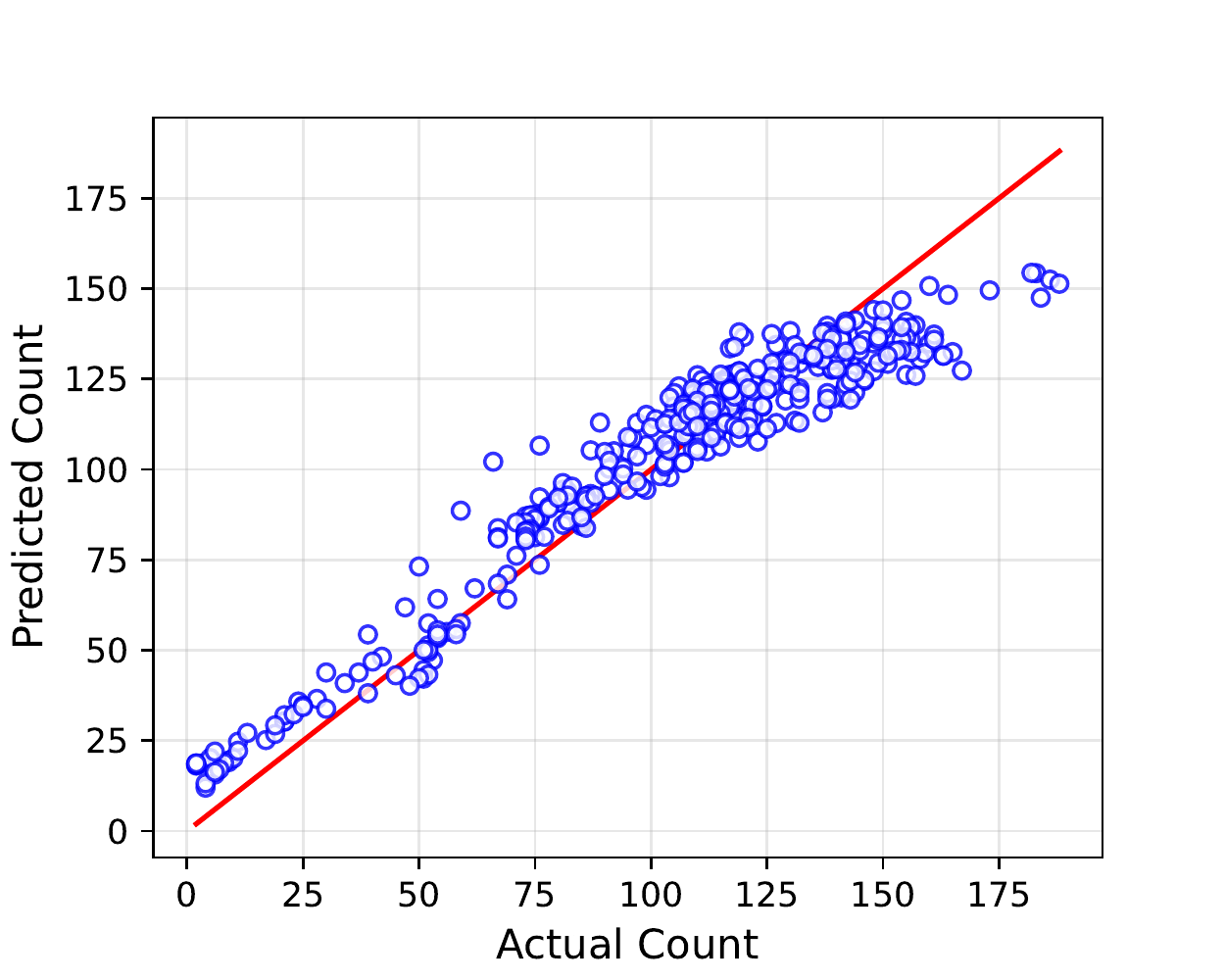}
    \caption{Counting errors in DroneNet over CARPK \cite{CARPK_dataset} dataset.}
    \label{fig:CARPK_errors}
\end{figure}

\begin{table}[!h]
\centering
\caption{Ablation Study on ShanghaiTech Part-B \cite{MCNN_CVPR2016} dataset.} \label{tab:ablation}
\begin{tabular}{cccccc} \toprule
&\multicolumn{2}{c}{ShanghaiTech Part-B} & &\multicolumn{2}{c}{CARPK} \\[0.8em] \cmidrule{2-3} \cmidrule{5-6}
Model &MAE &GAME & &MAE &GAME\\[0.5em] \midrule \midrule 
MCNN \cite{MCNN_CVPR2016}  &26.4  &55.2 &  &10.1 &43.4\\[0.5em]
DroneNet (ours)  &22.4  &41.9 &  &9.0 &40.1 \\[0.5em]
\bottomrule
\end{tabular}
\end{table}
The counting errors (i.e., actual count - predicted count) for the CARPK dataset \cite{CARPK_dataset} are plotted in Fig. \ref{fig:CARPK_errors}.

It can be observed that the model performs much better on a large number of images with low to medium densities. On high-density images ($< 150$ per image), the errors are relatively larger. High-density images are typically more challenging due to overlapping and occlusion effects.

\section{Conclusion} \label{sec:conclusion}
In this paper, we propose a novel deep learning model i.e., DroneNet using the Self-ONN learning paradigm instead of the commonly used CNN networks. DroneNet is lightweight, efficient, accurate, and is potentially a suitable choice for deployment over drones with limited computing resources on board. The performance of DroneNet is tested on benchmark datasets and the results report significant improvement over the equivalent CNN network (MCNN). Furthermore, the accuracy performance is very close to the CSRNet network exacerbates in terms of computational complexity and memory requirement. The DroneNet can run over edge devices much faster than CSRNet and almost at equal speed at equivalent deep architectures. In future work, we aim to test DroneNet on other benchmark datasets. We also aim to use Self-ONNs for designing dense crowd counting models for fine-grained counting in other applications.

\bibliographystyle{plainnat}
\bibliography{biblio}

\begin{thebibliography}{46}
\providecommand{\natexlab}[1]{#1}
\providecommand{\url}[1]{\texttt{#1}}
\expandafter\ifx\csname urlstyle\endcsname\relax
  \providecommand{\doi}[1]{doi: #1}\else
  \providecommand{\doi}{doi: \begingroup \urlstyle{rm}\Url}\fi

\bibitem[Aich and Stavness(2018)]{GSP_CVPR2019}
Shubhra Aich and Ian Stavness.
\newblock Global sum pooling: A generalization trick for object counting with
  small datasets of large images.
\newblock \emph{arXiv preprint arXiv:1805.11123}, 2018.

\bibitem[Allauddin et~al.(2019)Allauddin, Kiran, Kiran, Srinivas, Mouli, and
  Prasad]{Allauddin2019}
Md.~Saif Allauddin, G.~Sai Kiran, GSS.~Raj Kiran, G~Srinivas, G~Uma~Ratna
  Mouli, and P~Vishnu Prasad.
\newblock Development of a surveillance system for forest fire detection and
  monitoring using drones.
\newblock In \emph{IGARSS 2019 - 2019 IEEE International Geoscience and Remote
  Sensing Symposium}, pages 9361--9363, 2019.
\newblock \doi{10.1109/IGARSS.2019.8900436}.

\bibitem[Boominathan et~al.(2016)Boominathan, Kruthiventi, and
  Babu]{CrowdNet_CVPR2016}
Lokesh Boominathan, Srinivas S.~S. Kruthiventi, and R.~Venkatesh Babu.
\newblock Crowdnet: A deep convolutional network for dense crowd counting.
\newblock \emph{Proceedings of the 24th ACM international conference on
  Multimedia}, 2016.

\bibitem[Cao et~al.(2018)Cao, Wang, Zhao, and Su]{SANet_ECCV2018}
Xinkun Cao, Zhipeng Wang, Yanyun Zhao, and Fei Su.
\newblock Scale aggregation network for accurate and efficient crowd counting.
\newblock In \emph{ECCV}, 2018.

\bibitem[Chandana and Vasavi(2022)]{Chandana_2022}
V.~Sri Chandana and S.~Vasavi.
\newblock Autonomous drones based forest surveillance using faster r-cnn.
\newblock In \emph{2022 International Conference on Electronics and Renewable
  Systems (ICEARS)}, pages 1718--1723, 2022.
\newblock \doi{10.1109/ICEARS53579.2022.9752298}.

\bibitem[Chen et~al.(2012)Chen, Loy, Gong, and Xiang]{Chen_2012}
Ke~Chen, Chen~Change Loy, Shaogang Gong, and Tony Xiang.
\newblock Feature mining for localised crowd counting.
\newblock In \emph{BMVC}, 2012.

\bibitem[Dalal and Triggs(2005)]{Dalal_2005}
N.~Dalal and B.~Triggs.
\newblock Histograms of oriented gradients for human detection.
\newblock In \emph{2005 IEEE Computer Society Conference on Computer Vision and
  Pattern Recognition (CVPR'05)}, volume~1, pages 886--893 vol. 1, 2005.
\newblock \doi{10.1109/CVPR.2005.177}.

\bibitem[Davies et~al.(1995)Davies, Yin, and Velast{\'i}n]{Davies_1995}
Anthony~C. Davies, J.~H. Yin, and Sergio~A. Velast{\'i}n.
\newblock Crowd monitoring using image processing.
\newblock \emph{Electronics \& Communication Engineering Journal}, 7:\penalty0
  37--47, 1995.

\bibitem[Degerli et~al.(2021)Degerli, Zabihi, Kiranyaz, Hamid, Mazhar, Hamila,
  and Gabbouj]{hamila_2}
Aysen Degerli, Morteza Zabihi, Serkan Kiranyaz, Tahir Hamid, Rashid Mazhar,
  Ridha Hamila, and Moncef Gabbouj.
\newblock Early detection of myocardial infarction in low-quality
  echocardiography.
\newblock \emph{IEEE Access}, 9:\penalty0 34442--34453, 2021.

\bibitem[Felzenszwalb et~al.(2010)Felzenszwalb, Girshick, McAllester, and
  Ramanan]{Felzenszwalb_2010}
Pedro~F. Felzenszwalb, Ross~B. Girshick, David McAllester, and Deva Ramanan.
\newblock Object detection with discriminatively trained part-based models.
\newblock \emph{IEEE Transactions on Pattern Analysis and Machine
  Intelligence}, 32\penalty0 (9):\penalty0 1627--1645, 2010.
\newblock \doi{10.1109/TPAMI.2009.167}.

\bibitem[Gao et~al.(2019)Gao, Wang, and Gao]{MobileCount_PRCV2019}
Chenyu Gao, Peng Wang, and Ye~Gao.
\newblock Mobilecount: An efficient encoder-decoder framework for real-time
  crowd counting.
\newblock In \emph{Pattern Recognition and Computer Vision: Second Chinese
  Conference, PRCV 2019, Xi’an, China, November 8–11, 2019, Proceedings,
  Part II}, page 582–595. Springer-Verlag, 2019.

\bibitem[Gu and Lian(2022)]{MFCC_2022}
Siqi Gu and Zhichao Lian.
\newblock A unified multi-task learning framework of real-time drone
  supervision for crowd counting.
\newblock \emph{CoRR}, abs/2202.03843, 2022.
\newblock URL \url{https://arxiv.org/abs/2202.03843}.

\bibitem[Hamila et~al.(2022)Hamila, Ramanna, Henry, Kiranyaz, Hamila, Mazhar,
  and Hamid]{hamila_1}
Oumaima Hamila, Sheela Ramanna, Christopher~J. Henry, Serkan Kiranyaz, Ridha
  Hamila, Rashid Mazhar, and Tahir Hamid.
\newblock Fully automated 2d and 3d convolutional neural networks pipeline for
  video segmentation and myocardial infarction detection in echocardiography.
\newblock \emph{Multimedia Tools and Applications}, 2022.

\bibitem[He et~al.(2016)He, Zhang, Ren, and Sun]{ResNet_CVPR2016}
Kaiming He, X.~Zhang, Shaoqing Ren, and Jian Sun.
\newblock Deep residual learning for image recognition.
\newblock \emph{2016 IEEE Conference on Computer Vision and Pattern Recognition
  (CVPR)}, pages 770--778, 2016.

\bibitem[Hsieh et~al.(2017)Hsieh, Lin, and Hsu]{CARPK_dataset}
Meng-Ru Hsieh, Yen-Liang Lin, and Winston~H. Hsu.
\newblock Drone-based object counting by spatially regularized regional
  proposal network.
\newblock \emph{2017 IEEE International Conference on Computer Vision (ICCV)},
  pages 4165--4173, 2017.

\bibitem[Jiang et~al.(2019)Jiang, Xiao, Zhang, Zhen, Cao, Doermann, and
  Shao]{TEDnet_CVPR2019}
Xiaolong Jiang, Zehao Xiao, Baochang Zhang, Xiantong Zhen, Xianbin Cao,
  David~S. Doermann, and Ling Shao.
\newblock Crowd counting and density estimation by trellis encoder-decoder
  networks.
\newblock \emph{2019 IEEE/CVF Conference on Computer Vision and Pattern
  Recognition (CVPR)}, pages 6126--6135, 2019.

\bibitem[Khan et~al.(2022)Khan, Menouar, and Hamila]{Khan2022RevisitingCC}
Muhammad~Asif Khan, Hamid Menouar, and Ridha Hamila.
\newblock Revisiting crowd counting: State-of-the-art, trends, and future
  perspectives.
\newblock \emph{Image Vis. Comput.}, 129:\penalty0 104597, 2022.

\bibitem[Khan et~al.(2023)Khan, Hamila, Erbad, and Gabbouj]{khan_1}
Muhammad~Asif Khan, Ridha Hamila, Aiman Erbad, and Moncef Gabbouj.
\newblock Distributed inference in resource-constrained iot for real-time video
  surveillance.
\newblock \emph{IEEE Systems Journal}, 17\penalty0 (1):\penalty0 1512--1523,
  2023.

\bibitem[Kingma and Ba(2015)]{Adam_ICLR2015}
Diederik~P. Kingma and Jimmy Ba.
\newblock Adam: A method for stochastic optimization.
\newblock \emph{CoRR}, abs/1412.6980, 2015.

\bibitem[Kiranyaz et~al.(2020)Kiranyaz, Ince, Iosifidis, and Gabbouj]{ONN_2020}
Serkan Kiranyaz, Turker Ince, Alexandros Iosifidis, and M.~Gabbouj.
\newblock Operational neural networks.
\newblock \emph{Neural Computing and Applications}, 32:\penalty0 6645--6668,
  2020.

\bibitem[Li et~al.(2008)Li, Zhang, Huang, and Tan]{Li_2008}
Min Li, Zhaoxiang Zhang, Kaiqi Huang, and Tieniu Tan.
\newblock Estimating the number of people in crowded scenes by mid based
  foreground segmentation and head-shoulder detection.
\newblock In \emph{2008 19th International Conference on Pattern Recognition},
  pages 1--4, 2008.
\newblock \doi{10.1109/ICPR.2008.4761705}.

\bibitem[Li et~al.(2018)Li, Zhang, and Chen]{CSRNet_CVPR2018}
Yuhong Li, Xiaofan Zhang, and Deming Chen.
\newblock Csrnet: Dilated convolutional neural networks for understanding the
  highly congested scenes.
\newblock \emph{2018 IEEE/CVF Conference on Computer Vision and Pattern
  Recognition}, pages 1091--1100, 2018.

\bibitem[Lin and Davis(2010)]{Lin_2010}
Zhe Lin and Larry~S. Davis.
\newblock Shape-based human detection and segmentation via hierarchical
  part-template matching.
\newblock \emph{IEEE Transactions on Pattern Analysis and Machine
  Intelligence}, 32\penalty0 (4):\penalty0 604--618, 2010.
\newblock \doi{10.1109/TPAMI.2009.204}.

\bibitem[Liu et~al.(2019)Liu, Salzmann, and Fua]{CANNet_CVPR2019}
Weizhe Liu, Mathieu Salzmann, and Pascal~V. Fua.
\newblock Context-aware crowd counting.
\newblock \emph{2019 IEEE/CVF Conference on Computer Vision and Pattern
  Recognition (CVPR)}, pages 5094--5103, 2019.

\bibitem[Ma et~al.(2019)Ma, Wei, Hong, and Gong]{BL_ICCV2019}
Z.~Ma, X.~Wei, X.~Hong, and Y.~Gong.
\newblock Bayesian loss for crowd count estimation with point supervision.
\newblock In \emph{2019 IEEE/CVF International Conference on Computer Vision
  (ICCV)}, pages 6141--6150, Los Alamitos, CA, USA, nov 2019.

\bibitem[Malik et~al.(2021)Malik, Kiranyaz, and
  Gabbouj]{SelfONN_imgrestore_2021}
Junaid Malik, Serkan Kiranyaz, and M.~Gabbouj.
\newblock Self-organized operational neural networks for severe image
  restoration problems.
\newblock \emph{Neural networks : the official journal of the International
  Neural Network Society}, 135:\penalty0 201--211, 2021.

\bibitem[Peng et~al.(2020{\natexlab{a}})Peng, Li, and Zhu]{MMCNN_ACCV2020}
Tao Peng, Qing Li, and Peng~Fei Zhu.
\newblock Rgb-t crowd counting from drone: A benchmark and mmccn network.
\newblock In \emph{ACCV}, 2020{\natexlab{a}}.

\bibitem[Peng et~al.(2020{\natexlab{b}})Peng, Li, and Zhu]{DroneRGBT_dataset}
Tao Peng, Qing Li, and Pengfei Zhu.
\newblock Rgb-t crowd counting from drone: A benchmark and mmccn network.
\newblock In \emph{Computer Vision – ACCV 2020: 15th Asian Conference on
  Computer Vision, Kyoto, Japan, November 30 – December 4, 2020, Revised
  Selected Papers, Part VI}, page 497–513, Berlin, Heidelberg,
  2020{\natexlab{b}}. Springer-Verlag.

\bibitem[Ronneberger et~al.(2015)Ronneberger, Fischer, and Brox]{UNet_2015}
Olaf Ronneberger, Philipp Fischer, and Thomas Brox.
\newblock U-net: Convolutional networks for biomedical image segmentation.
\newblock \emph{ArXiv}, abs/1505.04597, 2015.

\bibitem[Sam et~al.(2017)Sam, Surya, and Babu]{SCNN_CVPR2017}
D.~Sam, S.~Surya, and R.~Babu.
\newblock Switching convolutional neural network for crowd counting.
\newblock In \emph{2017 IEEE Conference on Computer Vision and Pattern
  Recognition (CVPR)}, pages 4031--4039, Los Alamitos, CA, USA, jul 2017. IEEE
  Computer Society.

\bibitem[Sambolek and Ivasic-Kos(2021)]{Sambolek2021}
Sasa Sambolek and Marina Ivasic-Kos.
\newblock Automatic person detection in search and rescue operations using deep
  cnn detectors.
\newblock \emph{IEEE Access}, 9:\penalty0 37905--37922, 2021.
\newblock \doi{10.1109/ACCESS.2021.3063681}.

\bibitem[Simonyan and Zisserman(2015)]{VGG16_ICLR2015}
Karen Simonyan and Andrew Zisserman.
\newblock Very deep convolutional networks for large-scale image recognition.
\newblock In \emph{3rd International Conference on Learning Representations,
  {ICLR} 2015, San Diego, CA, USA, May 7-9, 2015, Conference Track
  Proceedings}, 2015.
\newblock URL \url{http://arxiv.org/abs/1409.1556}.

\bibitem[Simpson(2021)]{Simpson2021}
Todd Simpson.
\newblock Real-time drone surveillance system for violent crowd behavior
  unmanned aircraft system (uas) – human autonomy teaming (hat).
\newblock In \emph{2021 IEEE/AIAA 40th Digital Avionics Systems Conference
  (DASC)}, pages 1--9, 2021.
\newblock \doi{10.1109/DASC52595.2021.9594332}.

\bibitem[Sindagi and Patel(2017)]{CMTL_AVSS2017}
Vishwanath~A. Sindagi and Vishal~M. Patel.
\newblock Cnn-based cascaded multi-task learning of high-level prior and
  density estimation for crowd counting.
\newblock \emph{2017 14th IEEE International Conference on Advanced Video and
  Signal Based Surveillance (AVSS)}, pages 1--6, 2017.

\bibitem[Song et~al.(2021)Song, Wang, Wang, Tai, Wang, Li, Wu, and
  Ma]{SASNet_AAAI2021}
Qingyu Song, Changan Wang, Yabiao Wang, Ying Tai, Chengjie Wang, Jilin Li, Jian
  Wu, and Jiayi Ma.
\newblock To choose or to fuse? scale selection for crowd counting.
\newblock In \emph{AAAI}, 2021.

\bibitem[Sun et~al.(2022)Sun, Cao, Zhu, and Hu]{Sun2022}
Yiming Sun, Bing Cao, Pengfei Zhu, and Qinghua Hu.
\newblock Drone-based rgb-infrared cross-modality vehicle detection via
  uncertainty-aware learning.
\newblock \emph{IEEE Transactions on Circuits and Systems for Video
  Technology}, pages 1--1, 2022.
\newblock \doi{10.1109/TCSVT.2022.3168279}.

\bibitem[Szegedy et~al.(2015)Szegedy, Liu, Jia, Sermanet, Reed, Anguelov,
  Erhan, Vanhoucke, and Rabinovich]{Inception_CVPR2015}
Christian Szegedy, Wei Liu, Yangqing Jia, Pierre Sermanet, Scott~E. Reed,
  Dragomir Anguelov, D.~Erhan, Vincent Vanhoucke, and Andrew Rabinovich.
\newblock Going deeper with convolutions.
\newblock \emph{2015 IEEE Conference on Computer Vision and Pattern Recognition
  (CVPR)}, pages 1--9, 2015.

\bibitem[Tang et~al.(2022)Tang, Wang, and Chau]{TAFNET_2022}
Haihan Tang, Yi~Wang, and Lap-Pui Chau.
\newblock Tafnet: A three-stream adaptive fusion network for rgb-t crowd
  counting.
\newblock \emph{ArXiv}, abs/2202.08517, 2022.

\bibitem[Tian et~al.(2010)Tian, Sigal, Badino, la~Torre, and Liu]{Tian_2010}
Yan Tian, Leonid Sigal, Hern{\'a}n Badino, Fernando~De la~Torre, and Yong Liu.
\newblock Latent gaussian mixture regression for human pose estimation.
\newblock In \emph{ACCV}, 2010.

\bibitem[Topkaya et~al.(2014)Topkaya, Erdogan, and Porikli]{Topkaya_2014}
Ibrahim~Saygin Topkaya, Hakan Erdogan, and Fatih Porikli.
\newblock Counting people by clustering person detector outputs.
\newblock In \emph{2014 11th IEEE International Conference on Advanced Video
  and Signal Based Surveillance (AVSS)}, pages 313--318, 2014.
\newblock \doi{10.1109/AVSS.2014.6918687}.

\bibitem[Tuzel et~al.(2008)Tuzel, Porikli, and Meer]{Tuzel_2008}
Oncel Tuzel, Fatih Porikli, and Peter Meer.
\newblock Pedestrian detection via classification on riemannian manifolds.
\newblock \emph{IEEE Transactions on Pattern Analysis and Machine
  Intelligence}, 30\penalty0 (10):\penalty0 1713--1727, 2008.
\newblock \doi{10.1109/TPAMI.2008.75}.

\bibitem[Wang and Breckon(2022)]{SGANet_IEEEITS2022}
Qian Wang and T.~Breckon.
\newblock Crowd counting via segmentation guided attention networks and
  curriculum loss.
\newblock \emph{IEEE Transactions on Intelligent Transportation Systems}, 2022.

\bibitem[Wu and Nevatia(2006)]{Wu_2006}
Bo~Wu and Ramakant Nevatia.
\newblock Detection and tracking of multiple, partially occluded humans by
  bayesian combination of edgelet based part detectors.
\newblock \emph{International Journal of Computer Vision}, 75:\penalty0
  247--266, 2006.

\bibitem[Zeng et~al.(2017)Zeng, Xu, Cai, Qiu, and Zhang]{MSCNN_ICIP2017}
Lingke Zeng, Xiangmin Xu, Bolun Cai, Suo Qiu, and Tong Zhang.
\newblock Multi-scale convolutional neural networks for crowd counting.
\newblock \emph{2017 IEEE International Conference on Image Processing (ICIP)},
  pages 465--469, 2017.

\bibitem[Zhang et~al.(2015)Zhang, Li, Wang, and Yang]{CrowdCNN_CVPR2015}
Cong Zhang, Hongsheng Li, Xiaogang Wang, and Xiaokang Yang.
\newblock Cross-scene crowd counting via deep convolutional neural networks.
\newblock \emph{2015 IEEE Conference on Computer Vision and Pattern Recognition
  (CVPR)}, pages 833--841, 2015.

\bibitem[Zhang et~al.(2016)Zhang, Zhou, Chen, Gao, and Ma]{MCNN_CVPR2016}
Yingying Zhang, Desen Zhou, Siqin Chen, Shenghua Gao, and Yi~Ma.
\newblock Single-image crowd counting via multi-column convolutional neural
  network.
\newblock In \emph{2016 IEEE Conference on Computer Vision and Pattern
  Recognition (CVPR)}, pages 589--597, 2016.
\newblock \doi{10.1109/CVPR.2016.70}.

\end{thebibliography}







\end{document}